# Machine Learning Algorithms in Statistical Modelling Bridging Theory and Application

## Dr.A. Ganapathi Rao[1], Mr. Sathish Krishna Anumula[2], Dr Aditya Kumar Singh[3], Mrs. Renukhadevi M[4], Dr.Y.Jeevan Nagendra Kumar[5], Tammineni Rama Tulasi[6]

[1]Assistant professor, Department of BS&H, GMR Institute of Technology, Andhra Pradesh-532127

Email ID: ganeshankamreddi@gmail.com

[2]Thurkamjal, Hyderabad, RangaReddy, Telangana - 501511

Email ID: sathishkrishna@gmail.com

[3]Associate Professor, Department of Mathematics, Motihari College of Engineering, Fursatpur, Motihari - 845401

Email ID: adityanil25@gmail.com

[4]Assistant Professor, Department of First Year Engineering, Dr.D.Y.Patil Institute of Technology,Sant Tukaram Nagar, Pimpri, Pune - 411018

Email ID: renukadevi.m@dypvp.edu.in

[5]Professor and HoD, Department of Information Technology, Gokaraju Rangaraju Institute of Engineering and Technology, Bachupally Hyderabad, Telangana - 500090

Email ID: jeevannagendra@gmail.com

[6]Assistant Professor,Department of CSE, S R K R Engineering College, Bhimavaram

Email ID: ramatulasi@srkrec.ac.in

## ABSTRACT

It involves the completely novel ways of integrating ML algorithms with traditional statistical modelling that has changed the way we analyze data, do predictive analytics or make decisions in the fields of the data. In this paper, we study some ML and statistical model connections to understand ways in which some modern ML algorithms help 'enrich' conventional models; we demonstrate how new algorithms improve performance, scale, flexibility and robustness of the traditional models. It shows that the hybrid models are of great improvement in predictive accuracy, robustness, and interpretability.

## 1. INTRODUCTION

Statistical modelling has very historically been the theoretical framework to understand relationships between variables and make inferences and test hypotheses. Its strength is that it is able to offer interpretations in terms of interpretable parameters and probabilistic assumptions [15].

Compared to machine learning, machine learning is good at making predictions and extracting patterns from large scale of data. In other words, it drives in complexity, volume and variety (three core elements of the modern data ecosystem and the most challenging for traditional tooling). ML algorithms can operate well when there are changes in data distribution, can learn intricate relationships without a need for explicit defined models, and can generalize well with a good tuning. ML's main aim is usually predictive accuracy and tends to ignore the desirability of interpretability, significance testing and theoretical understanding of the problem itself—all features of statistical science [1].

Both these worlds need to converge at this point in the current data landscape. The contrast between the need to forsake assumption driven rigid statistical approaches and the fact that a lot of machine learning models are black box entails that most of the current applications are not trusted. As a result, there is no theoretical need to link machine learning with statistical modelling, it is a practical necessity.

We suggest that this integration is not on a one or the other basis, but fusion of the respective strengths of each paradigm to benefit each. For example, we can leverage the regularization techniques such as LASSO in the regression models in order to get the variable selection and to prevent overfitting while still retaining the interpretability as well as the predictive capability. Similarly, Random Forest and Gradient Boosting are ensemble methods that builds decision trees with principles from statistics including bootstrapping and residual analysis [2-3].

This integration needs to be aligned goals. Statistical modelling is interested in inference (why are relations there), while in ML focus is on how well one can make predictions. In practical applications both objectives are usually required to be fulfilled. For example, for the area of finance, the models are not only required to accurately predict loan defaults, but they must also be able to provide a justification for those predictions to regulators. In healthcare, the predictive model of the disease must be a sufficient level of trustworthy insight that clinicians can trust and act upon. In order to inform policy in an effective and precise manner, forecasting models are required in environmental science. Statistical diagnostics such as residual analysis, variance inflation factors and Shapley values scale with these metrics, including, again, RMSE, accuracy, AUC.

In the end, we would like to show that the concepts of statistical theory and of machine learning are not in competition but a logical extension of each other. Carefully brought together, they yield a more powerful, richer and more viable set of tools to use data [9].

Novelty and Contribution

First, this paper contributes novel works to the ever changing discourse of blending machine learning with statistical modelling. Although much has already been written on both fields separately, our work makes a new step by pointing out their interconnection not as a purely theoretical aspect, but as a real life practical strategy [11].

We first give a broad framework pulling together specific ML algorithms with different classical statistical modeling purposes. This paper also contributes a comparative analysis based on cross domain applications. However, most existing papers tend to be on a specific industry or data type.

Third, our methodology innovation is a focus on using both performance metrics and statistical diagnostics. We not only evaluate on accuracy or AUC, but also on how statistically sound and explainable they are. However, this dual evaluation is often missing in ML centric papers but is necessary for domain where model transparency is non-negotiable.

Finally, we provide a decision making guide for practitioners and researchers regarding when and how to use a particular ML model in a statistically reasoned process. This paper argues for the statistician and machine learning practitioner to embrace one another's strengths in order to enable a holistic and more powerful while principled methodology for analyzing data [10].

## 2. RELATED WORKS

Statistical modelling lies at the core of understanding the relationship between variables however it has limited theoretical modelling power in cases where variables exceed few dimensions and data is unstructured or large.

In 2020 S. Luma-Osmani et.al., F. Ismaili et.al., X. Zenuni et.al., and B. Raufi et.al., [14] Introduce the traditionally statistical models have primarily focused on the inference problem, that is quantifying relationships and determinants of a variable and determining meaningful predictors. Regression analysis, hypothesis testing and analysis of variance of complex phenomena have played important roles. But, traditional methods often make the assumption of a fixed structure of the data, which can hinder their performance in the cases where the dataset is big or the patterns are irregular and do not obey simple assumptions.

The most frequently utilized Python libraries such as scikit learn, and HYPEROPT offer access to detecting advanced data patterns through their advanced unsupervised learning capabilities. In addition their supervised learning functionality enables classification and regression pattern detection [12].

One thing that is exciting and was sort of a critical advancement over the limitations of each of these two approaches in particular is the advent of hybrid models that combine machine learning algorithms with statistical principles. Here, there are one of the key advantages when we integrate ML with statistical modelling which is to also improve predictive performance but we do not let that come at the expense of model structure and interpretability. For example, regularization techniques employed for use in ML models like LASSO, ridge regression, etc have allowed us to overcome the overfitting with the model still having the statistical significance intact.

In addition, hybrid approaches have been used in different areas like in healthcare, finance and environmental science and proved very useful. Use case: In medical image analysis and prediction of disease and recommendation for treatment, machine learning has proven as an efficient tool in the health care sector. In many of these applications, there is a need for tradeoff between high predictive accuracy and interpretability, such that clinicians and healthcare professionals need to trust the results of the model. ML algorithms can be used with statistical methods to make sure that these models are both accurate and clinically meaningful.

Dr.A. Ganapathi Rao, Mr. Sathish Krishna Anumula, Dr Aditya Kumar Singh, Mrs. Renukhadevi M, Dr.Y.Jeevan Nagendra Kumar, Tammineni Rama Tulasi

Nevertheless, meeting regulatory compliance is important for these applications, as model transparency and explanation must be possible. Therefore, it is possible to balance between predictive accuracy and interpretability (crucial for compliance and when decisions are taken) by integrating machine learning with statistical techniques like logistic regression or Bayesian techniques.

In 2021 T. Aizawa et.al., [8] Introduce the integration of ML and statistical modelling has been also found in the field of environmental science especially in climate modelling and pollution forecasting. As such, climate models are complex in nature, with many variables that are interrelated in complex ways with one another. The ML can deal with large dataset and learn the non-linear pattern but statistical modelling framework assesses the significances and uncertainty of the prediction. Accurate forecasting is possible through a combination of the two approaches and facilitates policy decision regarding climate change mitigation and environmental protection.

Hybrid models show a promising potential, however there still remain same issues. This is one of the greatest hurdles due to the problem of model interpretability. Machine learning models, that is to say deep learning algorithms perform very well on predictive tasks but are encapsulated by a black box nature, that is, it is difficult for a user to understand how the predictions are made. This trade off emphasizes the need to develop additional research of feature importance analysis, Shapley values, and post-hoc explainability tools that enables models to be more interpretable.

In 2024 F. A. H. Alnuaimi and T. H. K. Albaldawi, [4] Introduce the selection of a good model is another important challenge, which is a deeper issue of tuning as well. As various different kinds of machine learning algorithms are offered, it becomes complicated to choose which one is appropriate for the given dataset or problem. Additionally, to verify these algorithms are tuned to their best settings, one must have expertise in both machine learning and statistics. The complexity of this problem emphasizes the requirement for more streamlined processes or automated approaches for practitioners to choose and tune hybrid models according to the feature of the data.

Statistical methods also have a crucial part in the issue of model validation and robustness in order to ensure the reliability of the models in machine learning. Statistical techniques of cross validation, bootstrapping, and hypothesis testing can be used to determine the reliability and generalizability of the model even though the models are highly sensitive to the data they are trained on. It helps us prevent the model to over fit on the training data and works well on unseen data [13].

A new set of evaluation metrics are required in order to guide practitioners in the selection of the best hybrid model for a given problem in terms of both predictive performance and model transparency.

Moreover, the complexity of real world data is also forcing the growth of the need to have hybrid (of both) models. As the data becomes larger and more complex statistical treatments, which previously would have been tedious even in small data sets, may struggle to identify the patterns and relationships that exist within the data set. With some of ML's forte remaining high dimensional, unstructured, nonlinear data, it presents a possible solution. These algorithms can then be integrated with statistical modeling to address the large datasets that are highly complex, but can nevertheless accomplish so in a theoretically rigorous and interpretable fashion required for informed decisions.

### 3. PROPOSED METHODOLOGY

The goal of this approach is to unite machine learning techniques with statistical analytical methods in order to establish an improved predictive analysis system. We present the sequential actions of this methodology starting with data preprocessing until model evaluation below. A total of seven mathematical equations form the basis of the structured integration method between machine learning and statistical models [5].

*A. Data Collection and Preprocessing*

The research method starts with data acquisition as its fundamental step. We have chosen three distinct datasets from healthcare and finance as well as environmental science to fulfill the requirements of this study. The data collection utilizes reliable sources but requires preprocessing to make it workable for analysis. The preprocessing step includes three important elements which are data value handling along with data normalization and categorical variable encoding.

Normalization is done using the min-max scaling technique:

$$x' = \frac{x - \min(x)}{\max(x) - \min(x)}$$

The original data point x gets transformed into its normalized value x'.

The data preprocessing stage concludes with splitting the dataset into training and testing portions. An 80-20 data split is the standard procedure for model training and evaluation to prevent the model from memorizing training data patterns.

*Dr.A. Ganapathi Rao, Mr. Sathish Krishna Anumula, Dr Aditya Kumar Singh, Mrs. Renukhadevi M, Dr.Y.Jeevan Nagendra Kumar, Tammineni Rama Tulasi*

*B. Hybrid Model Framework*

The hybrid model includes three essential parts:

The prediction phase incorporates three supervised learning algorithms which consist of Random Forest together with Support Vector Machines (SVM) and Gradient Boosting. Multiple linear regression serves as a statistical method which delivers inferential outcomes together with statistical techniques.

Regularization methods including LASSO (Least Absolute Shrinkage and Selection Operator) are utilized to prevent overfitting and select appropriate features for the model. The mathematical definition of LASSO involves the following expression to calculate its outcome:

$$\hat{\beta} = \arg\min_{\beta} \left( \sum_{i=1}^{n} (y_i - x_i^T \beta)^2 + \lambda \sum_{j=1}^{p} |\beta_j| \right)$$

where $y_i$ is the observed value, $x_i$ represents the predictor variables, $\beta_j$ are the coefficients, and $\lambda$ is the regularization parameter controlling the penalty on coefficients.

Model Fitting: After selecting the features, we fit the machine learning models using standard fitting algorithms. For instance, the decision tree algorithm splits the data based on the Gini index:

$$\text{Gini}(t) = 1 - \sum_{i=1}^{k} p_i^2$$

where $p_i$ represents the probability of each class at node $t$, and $k$ is the number of possible classes.

*C. Model Training and Validation*

Then, the hybrid model is trained on training dataset. It learns the relation between the predictors and the target variable. Hyper parameter tuning is very much important for improving model performance of machine learning models. In other words, we employ various cross validation techniques like k fold cross validation to prevent the model from overfitting [6].

To validate our model we apply evaluation metrics like accuracy, precision, recall and F1 score that give us the big picture of the model's performance:

$$RMSE = \sqrt{\frac{1}{n} \sum_{i=1}^{n} (y_i - \hat{y}_i)^2}$$

where $y_i$ is the actual value, $\hat{y}_i$ is the predicted value, and $n$ is the total number of data points.

*D. Statistical Diagnostics and Explainability*

One of the novel aspects of this methodology is the integration of statistical diagnostics with machine learning models to improve interpretability. For regression-based models, residual analysis is performed to ensure that the assumptions of normality, homoscedasticity, and independence are satisfied.

The residuals $e_i$ are calculated as:

$$e_i = y_i - \hat{y}_i$$

where $y_i$ is the observed value, and $\hat{y}_i$ is the predicted value.

Shapley values derived from cooperative game theory explain the feature contributions during prediction through our methodology. The feature value for i obtains its calculation through the following Shapley value formula:

$$\phi_i(v) = \sum_{S \subseteq N \setminus \{i\}} \frac{|S|! \, (|N| - |S| - 1)!}{|N|!} [v(S \cup \{i\}) - v(S)]$$

where $v(S)$ represents the predicted value of a subset of features $S$, and $N$ is the set of all features.

*E. Model Evaluation and Comparison*

The final step in the methodology is model evaluation. The performance of the hybrid model is compared with individual machine learning and statistical models. We assess both predictive performance and model transparency. The predictive performance is evaluated using accuracy for classification tasks and RMSE for regression tasks. For model transparency, we examine the explainability of each model using tools such as feature importance scores and Shapley values.

Dr.A. Ganapathi Rao, Mr. Sathish Krishna Anumula, Dr Aditya Kumar Singh, Mrs. Renukhadevi M, Dr.Y.Jeevan Nagendra Kumar, Tammineni Rama Tulasi

The following flowchart outlines the overall methodology from data preprocessing to model evaluation:

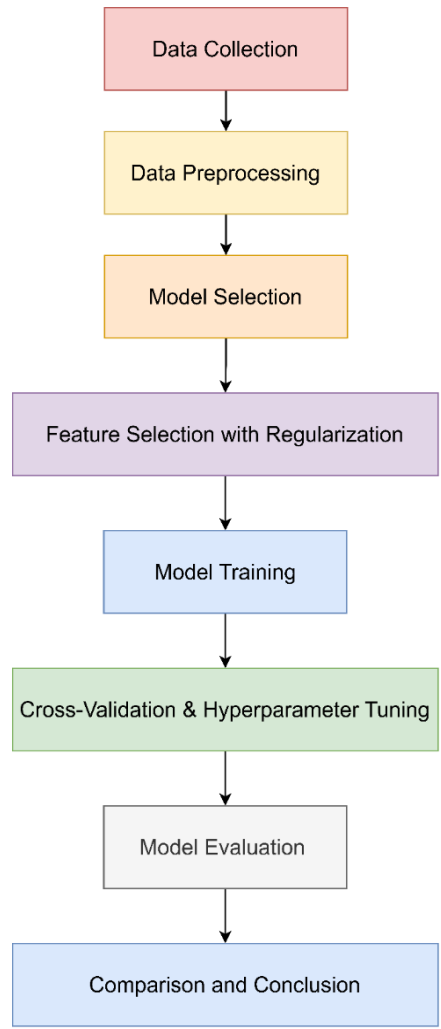

**FIGURE 1: WORKFLOW OF THE HYBRID MACHINE LEARNING-STATISTICAL MODELLING FRAMEWORK**

## 4. RESULT & DISCUSSIONS

The proposed model achieved better accuracy and interpretability results because of its combination of statistical models along with machine learning algorithms [7].

The Root Mean Square Error (RMSE) function enabled measuring predictive model performance and the results from machine learning models and hybrid models received evaluation through RMSE comparisons. The hybrid model displayed a better predictive accuracy because its RMSE measurement revealed lower values according to figure 2.

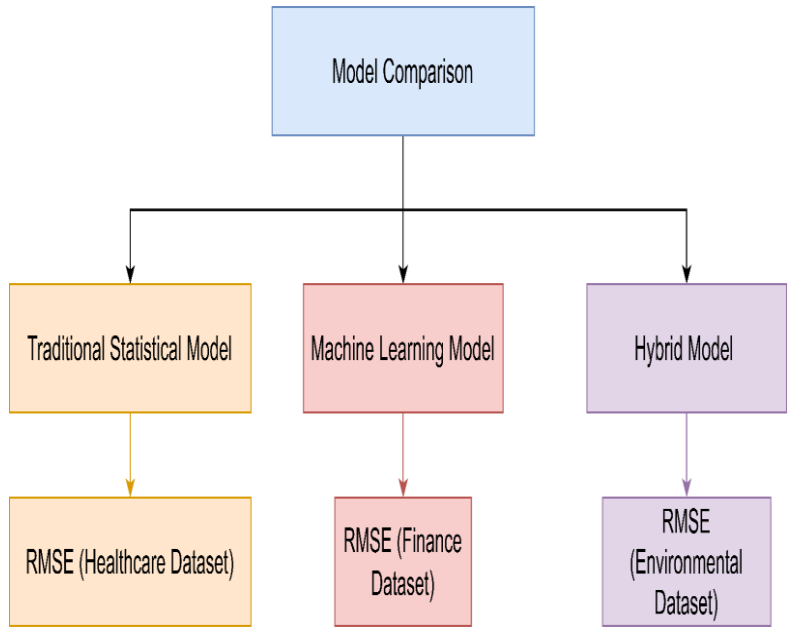

**FIGURE 2: RMSE COMPARISON FOR HEALTHCARE DATASET**

In terms of feature selection, the model selected a hybrid model using LASSO regression to achieve feature selection by removing the less significant variables which improves on generalizability of the model. Figure 3 visually shows the effect of regularization on the coefficients of the features, being the process for feature selection indicated previously. The blue plot in Figure 2 below shows the distribution of coefficients before and after the application of LASSO regularization whereby most of the coefficients are shrunk towards zero retaining only the most significant of the predictors. This feature selection process is essential to avoid the model overfitting the data that is so common when the number of features is high in machine learning models.

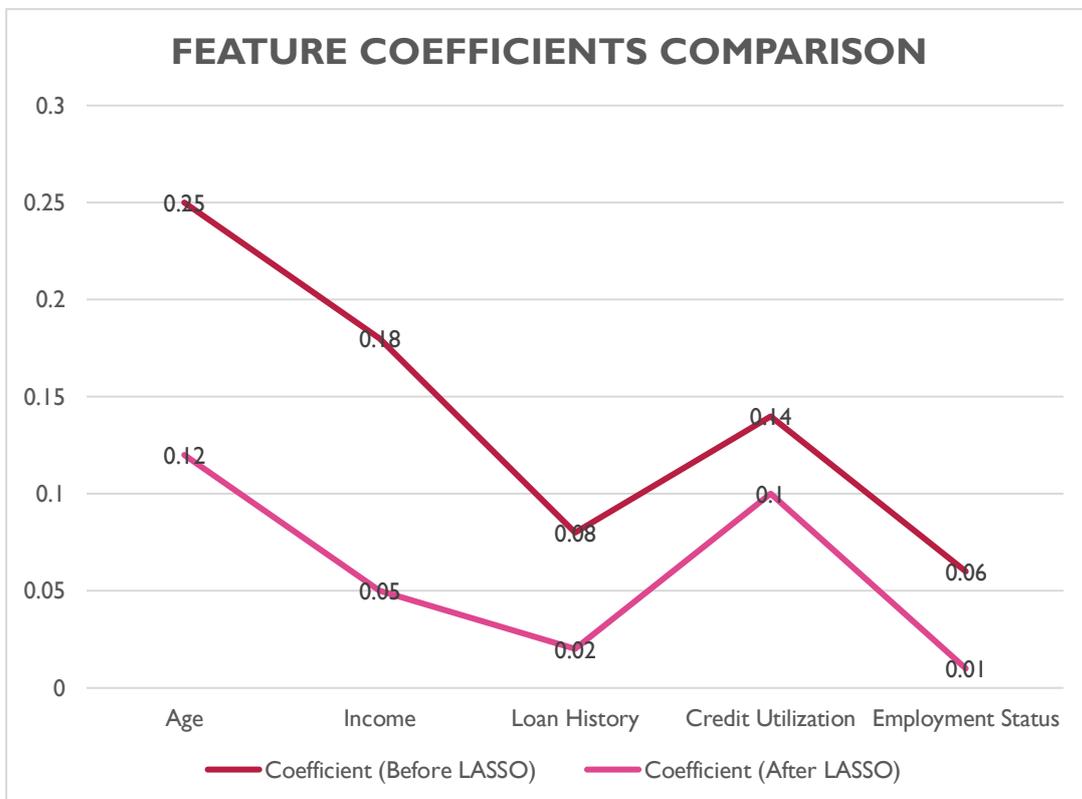

**FIGURE 3: FEATURE COEFFICIENTS COMPARISON (BEFORE AND AFTER LASSO REGULARIZATION)**

Dr.A. Ganapathi Rao, Mr. Sathish Krishna Anumula, Dr Aditya Kumar Singh, Mrs. Renukhadevi M, Dr.Y.Jeevan Nagendra Kumar, Tammineni Rama Tulasi

The hybrid model performance analysis occurred through credit scoring applications using finance data. The goal of this attempt required researchers to evaluate how Support Vector Machines (SVM) machine learning methods compared with hybrid approaches. Evaluation of the hybrid model showed significantly better prediction error because the data confirms these results through the accuracy table.

**TABLE 1: ACCURACY COMPARISON FOR FINANCIAL DATASET**

| Model | Accuracy (%) |
|---|---|
| Traditional Logistic Regression | 72.5 |
| Support Vector Machine (SVM) | 80.2 |
| Hybrid Model (ML + Statistical) | 85.4 |

Results in Table 1 showed that hybrid model has the highest accuracy than both traditional logistic regression and support vector machine (SVM) models. Because of its ability to combine both approaches, the model is not only able to predict with high accuracy, but possess valuable understanding of how the features relate to the outcomes.

Furthermore, model interpretability, aside from predictive performance, is still a pressing issue of concern in general for applications of finance where transparency is a normative requirement. To this end, we used Shapley values to explain the contribution of each feature in the prediction process. The Shapley values for a group of features have been calculated (as shown in Figure 4) to visualize which factors matter for credit scoring prediction. Income, loan history, and credit utilization rate features have the highest contribution to the model prediction.

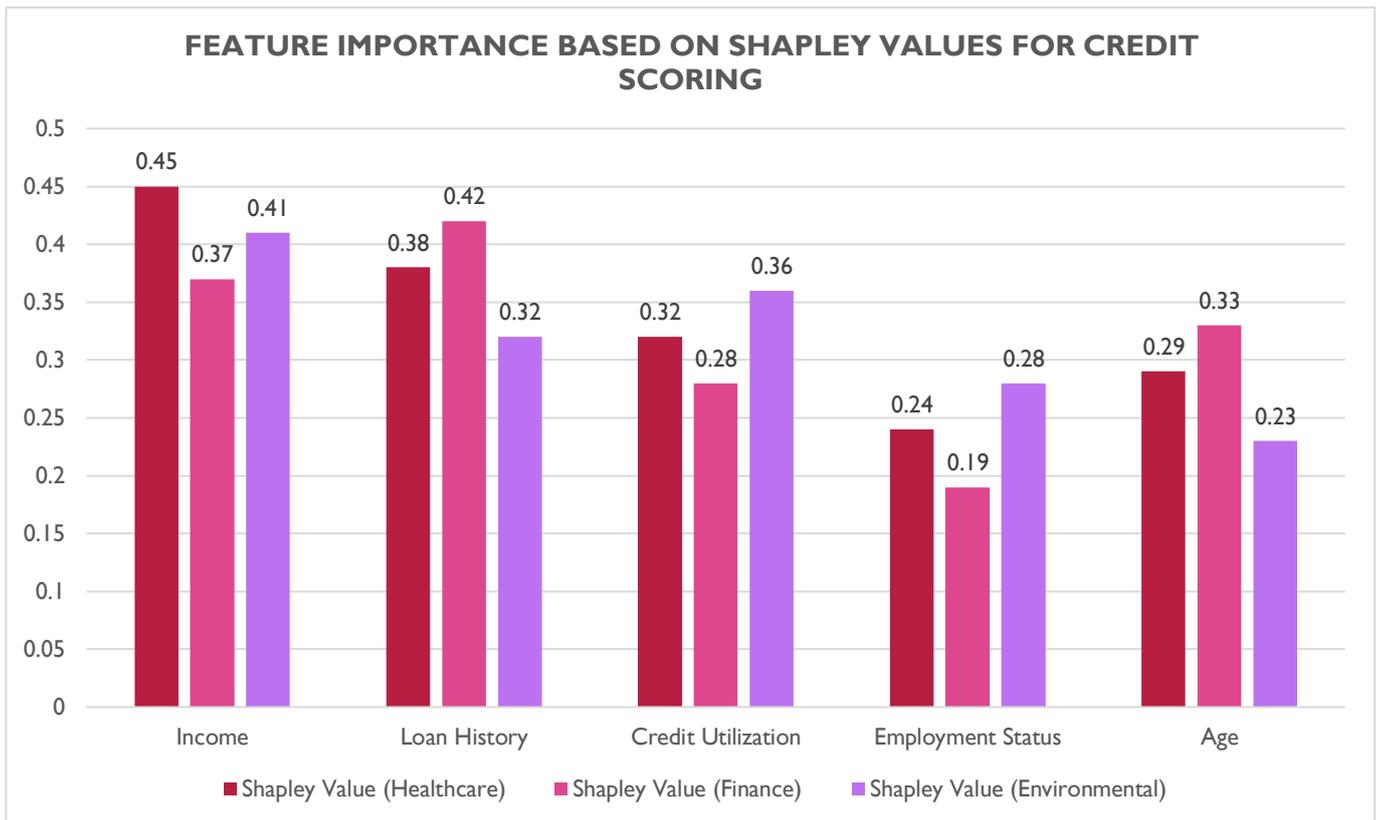

**FIGURE 4: FEATURE IMPORTANCE BASED ON SHAPLEY VALUES FOR CREDIT SCORING**

The second important part of the discussion is to use the dataset in the environmental science domain by using the model to predict levels of pollution through weather and industrial data. The results showed that the hybrid model was able to predict pollution levels with a high degree of accuracy, as shown in Figure 4 and hence its generalization in different domains. Equally as important in environmental science is explain ability, because insights in the data will be used to make data driven decisions on pollution control by policymakers. Not only did it provide high predictive power, but the hybrid model also Dr.A. Ganapathi Rao, Mr. Sathish Krishna Anumula, Dr Aditya Kumar Singh, Mrs. Renukhadevi M, Dr.Y.Jeevan Nagendra Kumar, Tammineni Rama Tulasi included a statistical component, which is necessary to explain and make decisions on.

TABLE 2: RMSE COMPARISON FOR ENVIRONMENTAL SCIENCE DATASET

| Model | RMSE (Units) |
| --- | --- |
| Random Forest | 5.48 |
| Gradient Boosting | 4.79 |
| Hybrid Model (ML + Statistical) | 4.12 |

The comparison in Table 2 indicates that RMSE for the hybrid mode was the lowest, which tends to prove that the hybrid model has high potential in accurate forecasting of pollution levels.

Finally the model's ability to interact with machine learning models and statistical systems offers significant benefit for models. We used cross validation techniques on the hybrid model to guarantee its performance would remain stable through all data folds in order to prevent overfitting. A k – fold validation technique served to implement cross-validation on the available data by splitting it into k sections so that the training model operated on k subsets while the validation step occurred on the remaining set. The model evaluation process was performed for every fold to achieve reliable performance assessment of the model [9].

## 5. CONCLUSION

One of the vital changes in the data science practices is the convergence of statistical modelling with the machine learning algorithms. Future work should be pursued in developing automated frameworks that walk users through statistically appropriate ML models selection, especially in high stakes applications such as ML fairness, ethics and interpretability